\documentclass{IOS-Book-Article}

\usepackage{mathptmx}
\usepackage{soul}\setuldepth{article}
\usepackage{amsmath}
\usepackage{algorithmic}
\usepackage[utf8]{inputenc}
\usepackage[english]{babel}
\usepackage{graphicx}
\usepackage{subcaption}
\usepackage{flafter}
\usepackage{float}
\usepackage{wrapfig}
\usepackage{tabularx}
\graphicspath{ {./} }

\def\hb{\hbox to 10.7 cm{}}
\begin{document}
	
	\pagestyle{headings}
	\def\thepage{}
	
	\begin{frontmatter}              

\title{Visualising Argumentation Graphs with Graph Embeddings and t-SNE}
\markboth{}{August 2020\hb}

\author[A]{\fnms{Lars} \snm{Malmqvist}%
	\thanks{Corresponding Author: Lars Malmqvist, University of York, E-mail:
		lm1775@york.ac.uk.}},
\author[A]{\fnms{Tommy} \snm{Yuan}}
\author[B]{\fnms{Suresh} \snm{Manandhar}}
\runningauthor{Malmqvist et al.}
\address[A]{University of York, UK}
\address[B]{Naamii, Kathmandu, Nepal}

\begin{abstract}
. This paper applies t-SNE, a visualisation technique familiar from Deep Neural Network research to argumentation graphs by applying it to the output of graph embeddings generated using several different methods. It shows that such a visualisation approach can work for argumentation and show interesting structural properties of argumentation graphs, opening up paths for further research in the area.
\end{abstract}

\end{frontmatter}
\markboth{August 2020\hb}{August 2020\hb}

\section{Introduction}
This paper examines whether using graph embeddings\cite{Goyal2018} with a dimensionality reducing algorithm, t-SNE\cite{vandermaaten2208}, to visualise argumentation graphs and their structural properties through unsupervised learning can lead to interesting results. This approach has been effective at data visualisation in Deep Neural Network research for graph-based methods\cite{Pezzotti2019}. The contribution of this paper is to show that this technique also holds promise in the visualisation of argumentation graphs by applying it at both the node and the graph level using both a standard and a custom built embedding approach. In particular, we show that it is possible to clearly visualise the functional partitions of arguments in the Sembuster domain\cite{Rodrigues2017} using this method and to separate argumentation domains into visual clusters at the graph level using a custom GCN embedding, raising the possibility that both differences between argument graphs and the function of arguments within argumentation graphs can be clustered and shown in a visually intuitive way using unsupervised methods.

\section{Graph Embeddings}
Graph Embeddings are used in a range of graph analysis applications including node classification, link prediction, clustering, and visualisation either directly or as additional input features to machine learning algorithms. There are several different algorithmic approaches to generating graph embeddings that are suited to different use cases. Goyal and Ferrara define three main approaches in their 2018 survey of the field\cite{Goyal2018}. 

The first category is factorisation-based approaches that share the property of working with a matrix representation (e.g. adjacency or Laplacian matrix) and a proximity measure to calculate the node embeddings. The next category is based on random walks through the graph to generate the embedding. Deepwalk, for instance, maximises the probability of seeing the last \textit{k} and the next \textit{k} nodes in the random walk centred at a given node. The last major category of approaches is based on Deep Neural Networks. For instance, Graph Convolutional Networks (GCN) generate an embedding by iteratively aggregating neighbourhood embeddings\cite{Kipf2016}. In this paper, we will use HOPE\cite{Ou2016}, a factorisation based approach, and a GCN based neural network approach to obtain the embeddings that we will visualise for argumentation graphs.

\section{Visualising Embeddings using t-SNE}
Embeddings are usually of low dimensionality, but even so, they are not easy to visualise. However, once an embedding has been generated for complex data types such as graphs, words, or images, it becomes possible to use dimensionality reduction techniques to visualise them effectively. 

In the Deep Neural Network community, the dimensionality reduction technique of choice is t-distributed stochastic neighbor embedding (t-SNE). In contrast to other common dimensionality reduction techniques such as Principal Component Analysis (PCA), t-SNE does not rely on a linear projection but uses local relationships between points. It uses Student's t-distribution to model the relationship between points in the higher dimensional space and then recreates those relationships in the lower dimensional space by way of a gradient descent based algorithm\cite{vandermaaten2208}.

A simple example that demonstrates the way this visualisation approach works can be found by applying it to the MNIST dataset of handwritten digits\cite{lecun-mnisthandwrittendigit-2010}\footnotetext[1]{Reproduced from \cite{Pezzotti2019}}. We see that t-SNE 2-D embeddings are able to cluster the 10 digits from MNIST dataset into 10 distinct clusters \cite{Pezzotti2019}. This shows that graph embeddings can extract visual similarity from the raw data without supervision.

\begin{figure}[h!]
	\centering
	\begin{subfigure}[b]{0.3\linewidth}
		\includegraphics[width=\linewidth]{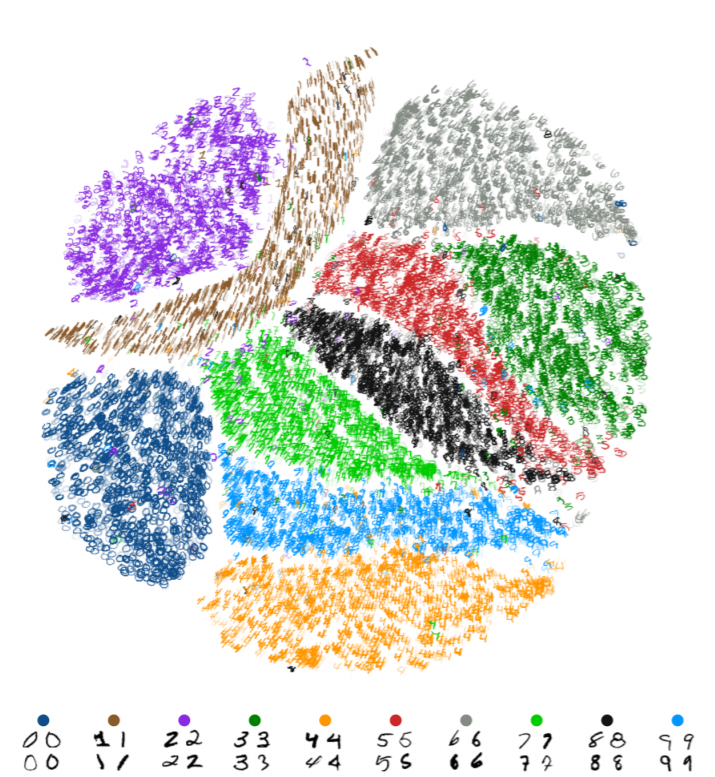}
		\caption{MNIST Visualised with t-SNE.}
	\end{subfigure}
	\begin{subfigure}[b]{0.3\linewidth}
		\includegraphics[width=\linewidth]{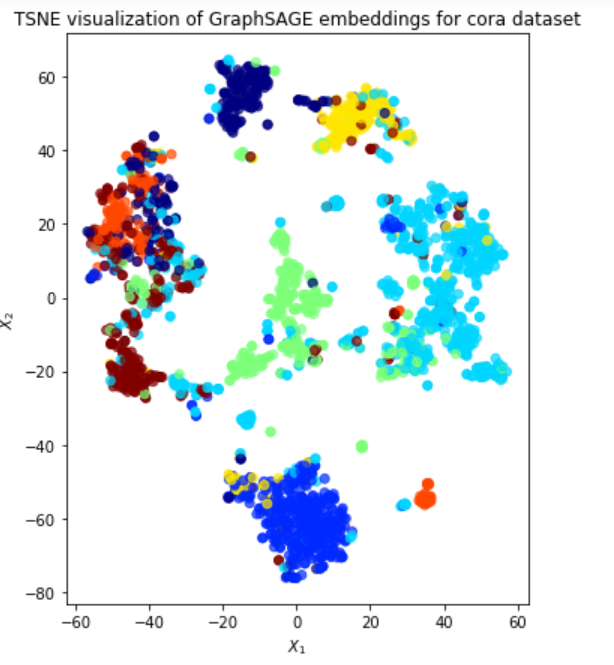}
		\caption{CORA Visualised with t-SNE.}
	\end{subfigure}
	\par
	\begin{subfigure}[b]{0.3\linewidth}
		\includegraphics[width=\linewidth]{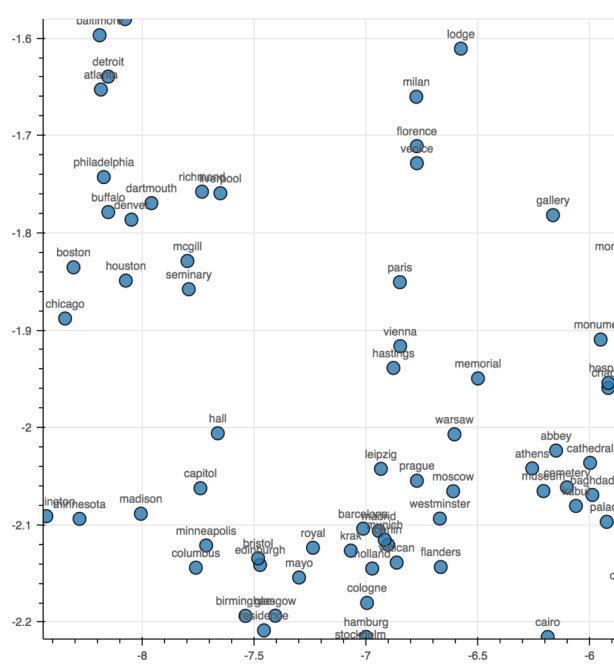}
		\caption{Word2Vec Visualised with t-SNE.}
	\end{subfigure}
	\caption{Examples of t-SNE Visualisation}
	\label{fig:mnist_cora2}	
\end{figure}

A more practical example can be found when using t-SNE with word embeddings. Word embeddings project words into an embedding space in an analogous way to graph embeddings. A good word embedding would, therefore, place words with similar meanings closer to each other in the embedding space. Using Word2Vec\cite{Mikolov2013}, a word embedding trained by trying to predict individual words given a context, the following chart of cities was generated. In general, cities from the same countries are placed closer together, although there are some anomalies\footnotetext[2]{Reproduced from https://nlpforhackers.io/word-embeddings/}. 

This approach also works well for graph-structured data. Once a graph embedding has been created for a graph it can also be visualised using t-SNE and the results will also tend to preserve a visual representation of the proximity measure that the embedding optimises. An example of this can be found by applying the GraphSAGE mode\cite{Hamilton2017}, a Deep Learning approach, to the CORA dataset of academic papers and citations\footnotetext[3]{Reproduced from https://towardsdatascience.com/using-graphsage-to-learn-paper-embeddings-in-cora-a94bb1e9dc9d}. This results in clusters on the class of the papers in the dataset, represented by the point colours, but with several exceptions that invite further investigation. In this case, we see that the information in the graph embeddings enables a clustering close to the underlying distribution of document classes without prior knowledge.

\section{Visualising Argumentation Graphs}
\subsection{Introduction}
Argumentation graphs are to some extent themselves a way of visualising arguments. However, for large argumentation structures with hundreds or thousands of arguments looking at the unadorned graph is of little utility. In the following sections, we will give two examples of how one can use graph embeddings to visualise information about the structure of arguments, one using node-level information to analyse a single argument and one to visualise properties of whole argumentation graphs. Both examples are based on the Abstract Argumentation formalism\cite{Dung1995}, but there is nothing inherent in this approach that limits it to one type of graph-based representation.

\subsection{Node-Level Visualisation}
To examine the possibilities of using Graph Embeddings and t-SNE to visualise argumentation graphs, we have started with a formal domain, Sembuster, that has the interesting property of having three types of arguments with distinct structural characteristics. The Sembuster domain, originally proposed by Caminada and Verheij\cite{Caminada2010}, is composed of unique graphs generated for each cardinality, $k$, that can be partitioned into three different types: A, B, and C. 
\begin{wrapfigure}{r}{0.4\linewidth}
	\begin{center}
		\includegraphics[width=\linewidth]{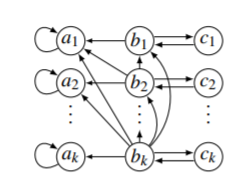}
	\end{center}
	\caption{Sembuster Scheme}
\end{wrapfigure}
Arguments in the A partition only attack themselves. An argument $B_i$ attacks all arguments $A_j$ where $i  >= j$, arguments $B_j$ where $i  > j$ and the argument $C_i$. An argument $C_i$ attacks the corresponding argument $B_i$. 

To visualise this configuration, we trained a graph embedding using the HOPE\cite{Ou2016} method on a set of Sembuster graphs with a cardinality from 300 to 4500. HOPE is a directionality preserving embedding that works well on directed graphs such as argumentation graphs. The graphs were trained with a dimensionality of 128 features using a single K80 GPU. Training time was from 48 seconds to 5 minutes 12 seconds for a single embedding. These graphs were then visualised using 2-dimensional t-SNE using a colour coding corresponding to the argument partitions and subjected to visual inspection. 
\begin{figure}[h!]
	\centering
	\begin{subfigure}[b]{0.3\linewidth}
		\includegraphics[width=\linewidth]{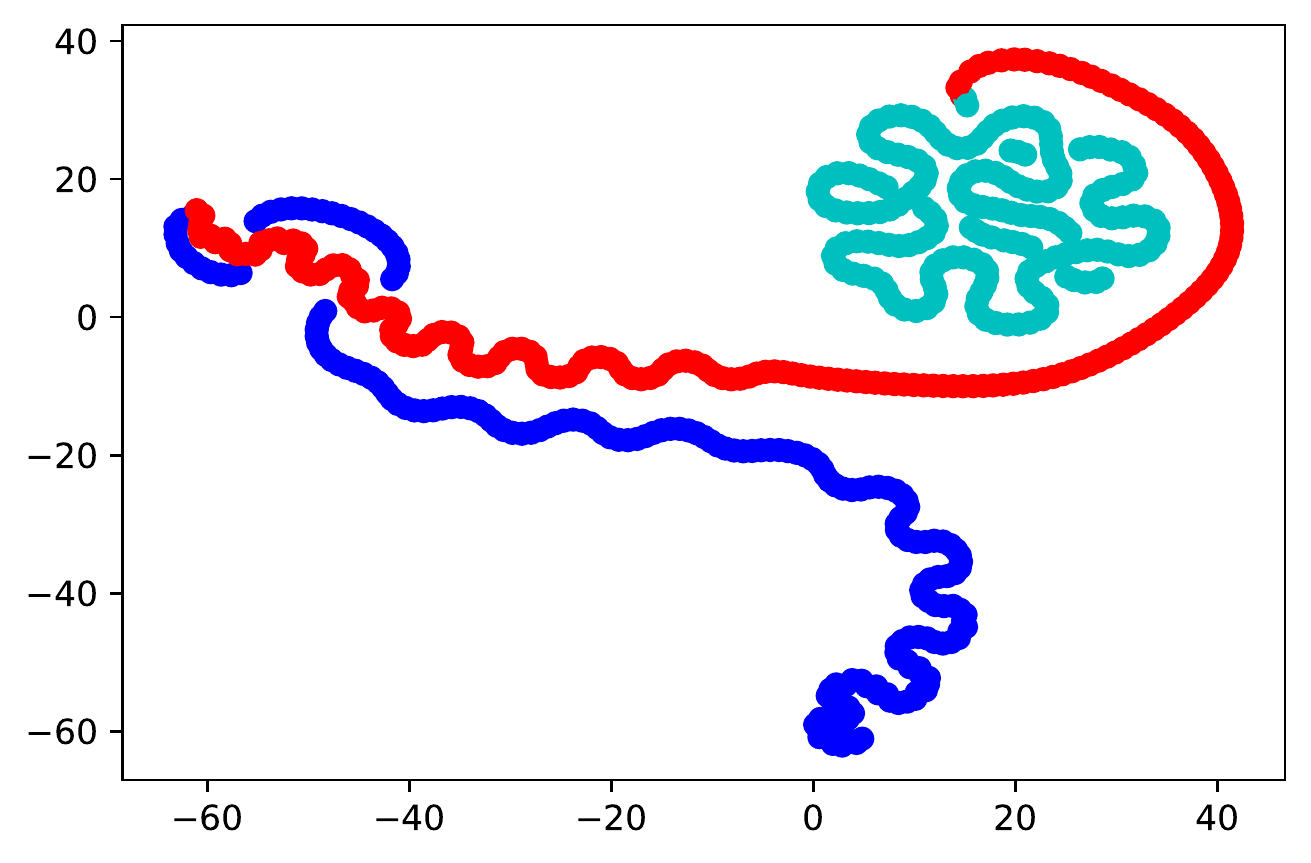}
		\caption{k=600}
	\end{subfigure}
	\begin{subfigure}[b]{0.3\linewidth}
		\includegraphics[width=\linewidth]{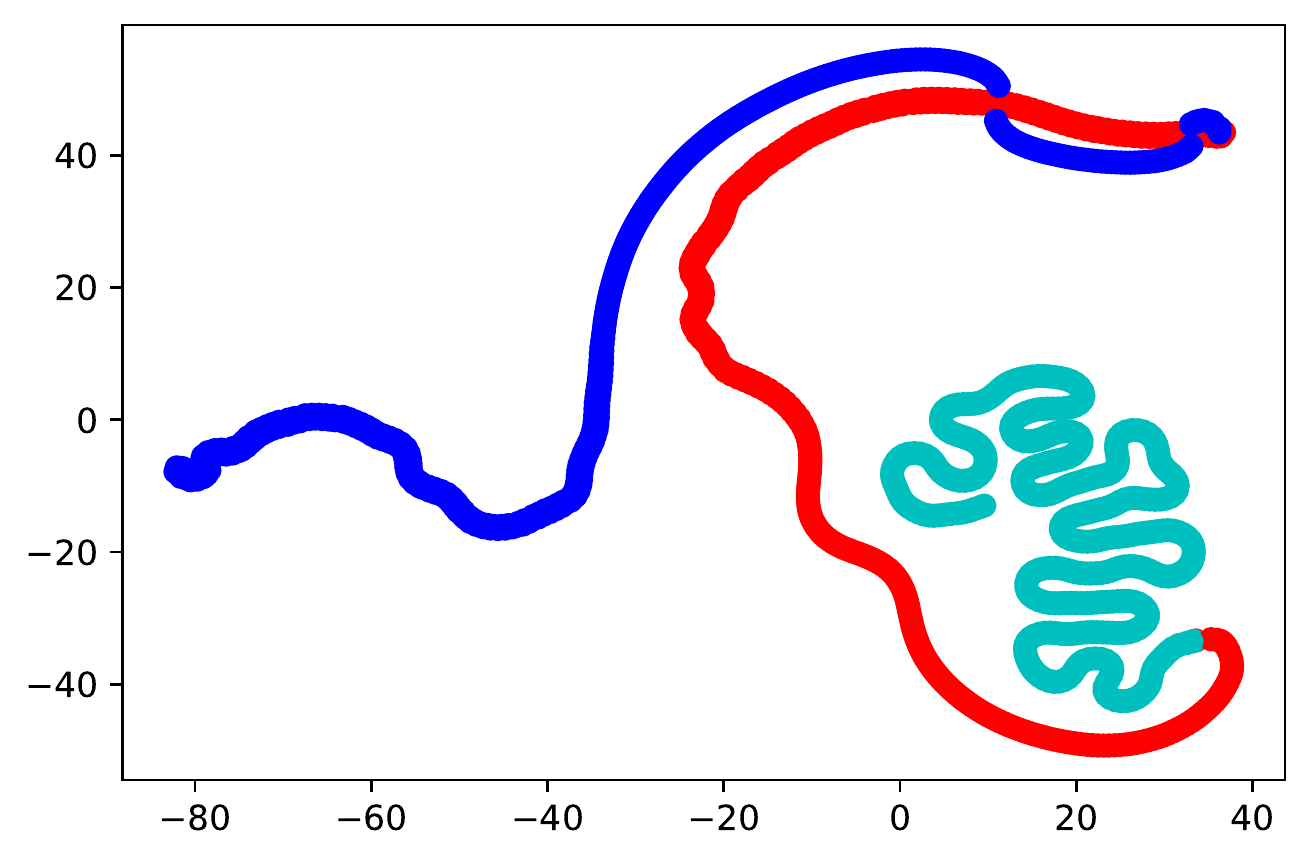}
		\caption{k=1200}
	\end{subfigure}
	\begin{subfigure}[b]{0.3\linewidth}
		\includegraphics[width=\linewidth]{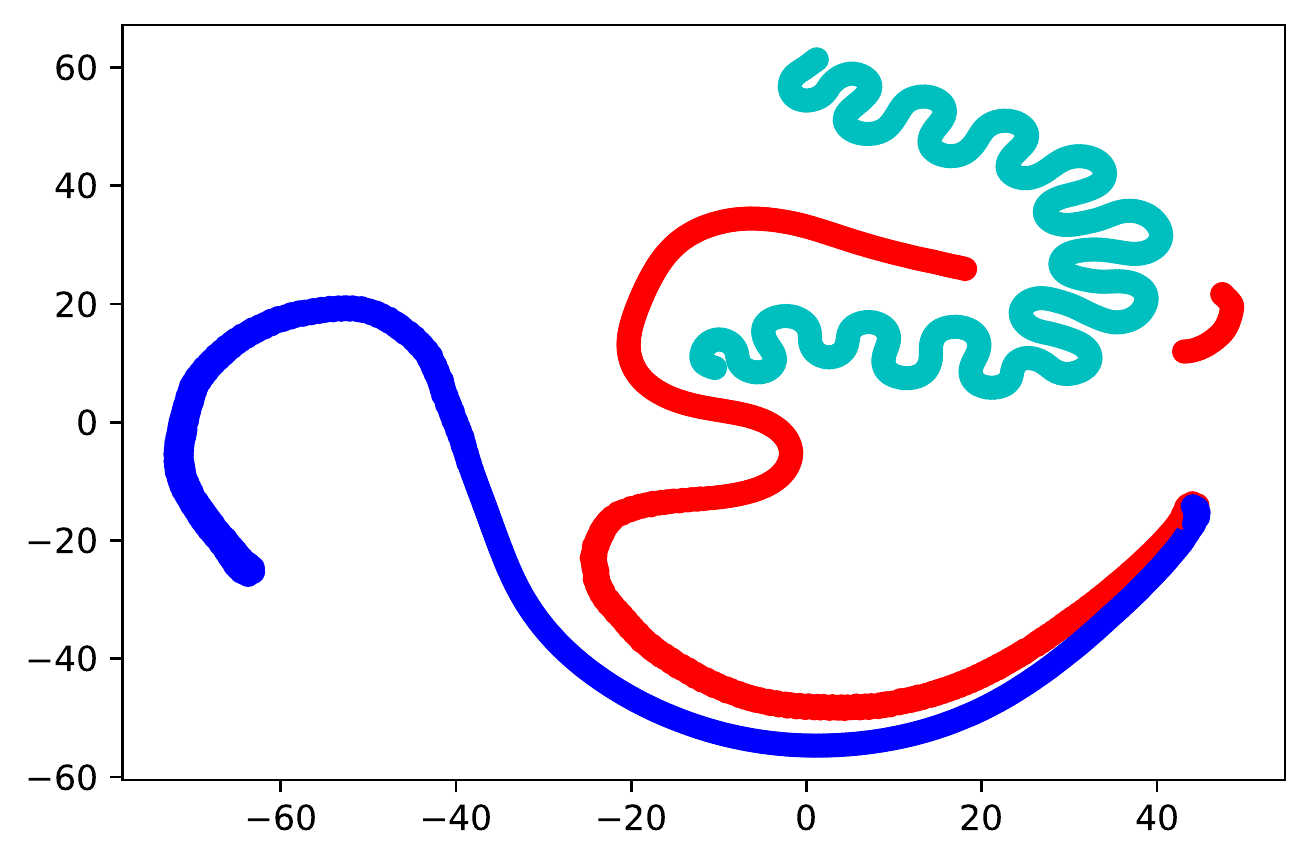}
		\caption{k=1800}
	\end{subfigure}
	\caption{T-SNE Visualisation of Sembuster Graphs. Cyan represents partition A, Red partition B, and Blue partition C.}
	\label{fig:mnist_cora}	
\end{figure}
If we examine the three examples, the three types of arguments stand out clearly. The self-referential A arguments are curled in on themselves in the visualisation and the related B arguments are closer to the A than the unrelated C arguments. 
The string-like construction of the Sembuster graphs seems at least superficially to also be represented in the string-like nature of the visualisation. While this is a highly formal example, it does indicate that the functional structure of arguments can in some cases be visualised using this type of technique, although its application to arguments of a less formal nature would need further investigation. Applying similar techniques on other argumentation graphs that are under analysis would potentially be able to show functional clusterings of arguments based on the type of graph embedding that has been applied. It may even be possible to design specific embedding approaches that allow training based on the particular properties we are interested in analysing for.

\subsection{Graph-Level Visualisation}
We also trained a GCN model\cite{Kipf2016} to classify variants of graphs from different abstract argumentation graph domains. We trained the embeddings on 10 separate domains using a custom built model. The domains are defined in Rodrigues et al.\cite{Rodrigues2017} and the table below give a description of their basis. The dataset was based on a subset of the tasks from the 2nd International Competition on Computational Models of Argumentation using an even sample drawn from these 10 domains, which all form part of the competition corpus. 

The GCN model used 4 convolutional layers followed by 2 fully connected layers to try to predict the class of a given input graph. The implication would be that the visualisation should show the graphs in a given domain following in a recognisable pattern (e.g. clustering or equal spacing). This could for instance be useful in situations where you have an argumentation corpus that is not clustered or classified and you want to group them by similarity and be able to show that similarity in a visual mode. After the model had been trained for 4 hours, the training was stopped and the output of the last convolutional layer was used as a graph embedding for visualisation by extracting the raw features of the last convolutional layer.

\begin{table}[H]
	\caption{Domains for Graph-Level Visualisation}
	\begin{tabularx}{\textwidth}{|l|l|X|}
		\textbf{Identifier} & \textbf{Domain}  & \textbf{Description}\\ 	
		afinput & ABA2AF & Assumption-Based Argumentation translated to abstract argumentation frameworks \\ 
		admbuster & AdmBuster & AdmBuster graphs, based on Caminada and Podlaszewski\cite{Caminada2017} \\ 
		BA & Barabasi-Albert & Barabasi-Albert graphs, randomly generated \\ 
		ER & Erdös-Rényi & Erdös-Rényi graphs, randomly generated  \\ 
		grd & GroundedGenerator & Randomly generated argumentation frameworks containing only a grounded extension \\ 
		.cnf & Planning2AF & Planning problems transformed to abstract argumentation problems \\ 
		sembuster & SemBuster & SemBuster graphs, see section 4.2\\ 
		scc & SccGenerator & Randomly generated argumentation frameworks containing multiplestrongly connected components  \\ 
		.gml. & Traffic  & Traffic networks converted to abstract argumentation frameworks\\ 
		WS & Watts-Strogatz & Watts-Strogatz graphs, randomly generated\\ 
	\end{tabularx}
	\label{tab:addlabel}%
\end{table}

\begin{figure}[h!]
	\centering
	\begin{subfigure}[b]{0.45\linewidth}
		\includegraphics[width=\linewidth]{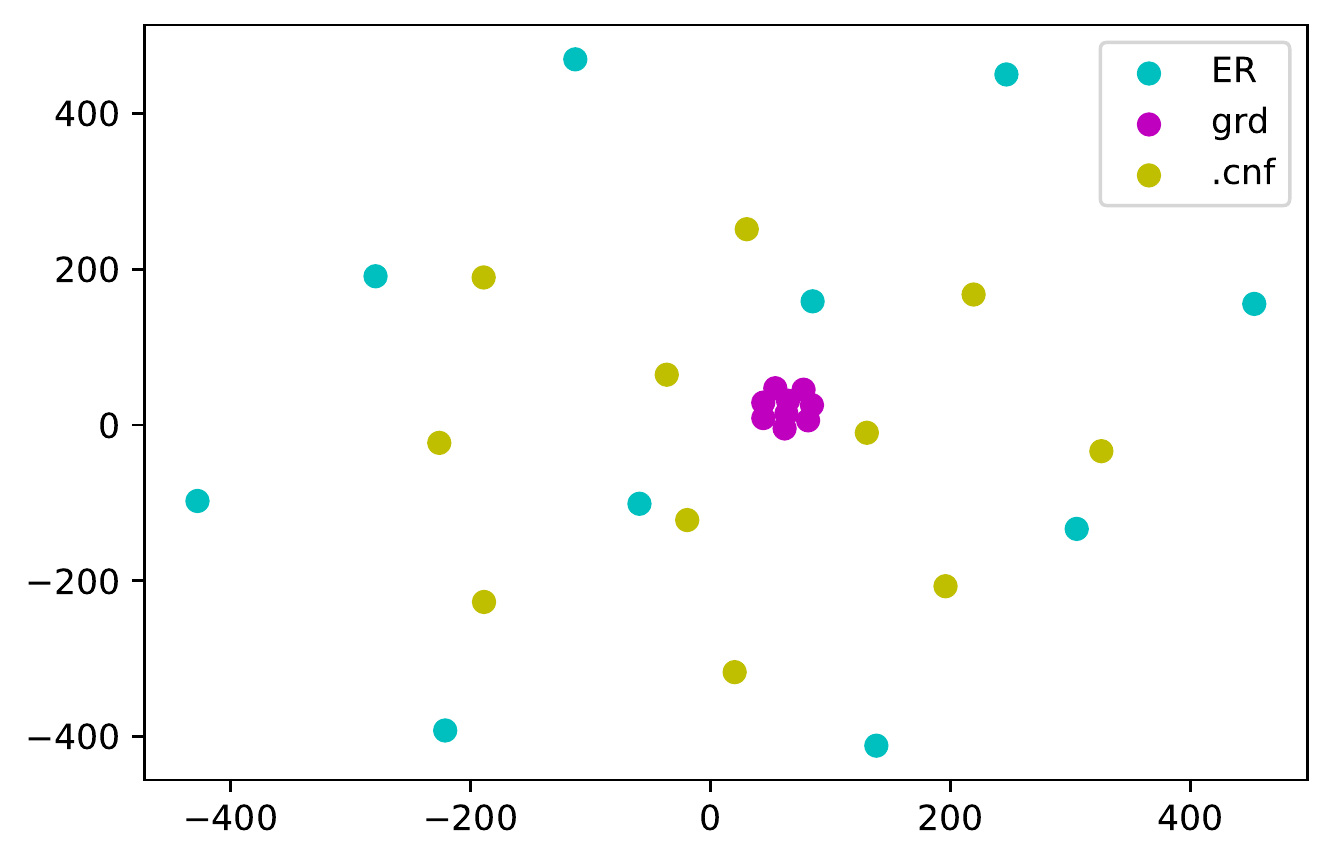}
	\end{subfigure}
	\begin{subfigure}[b]{0.45\linewidth}
		\includegraphics[width=\linewidth]{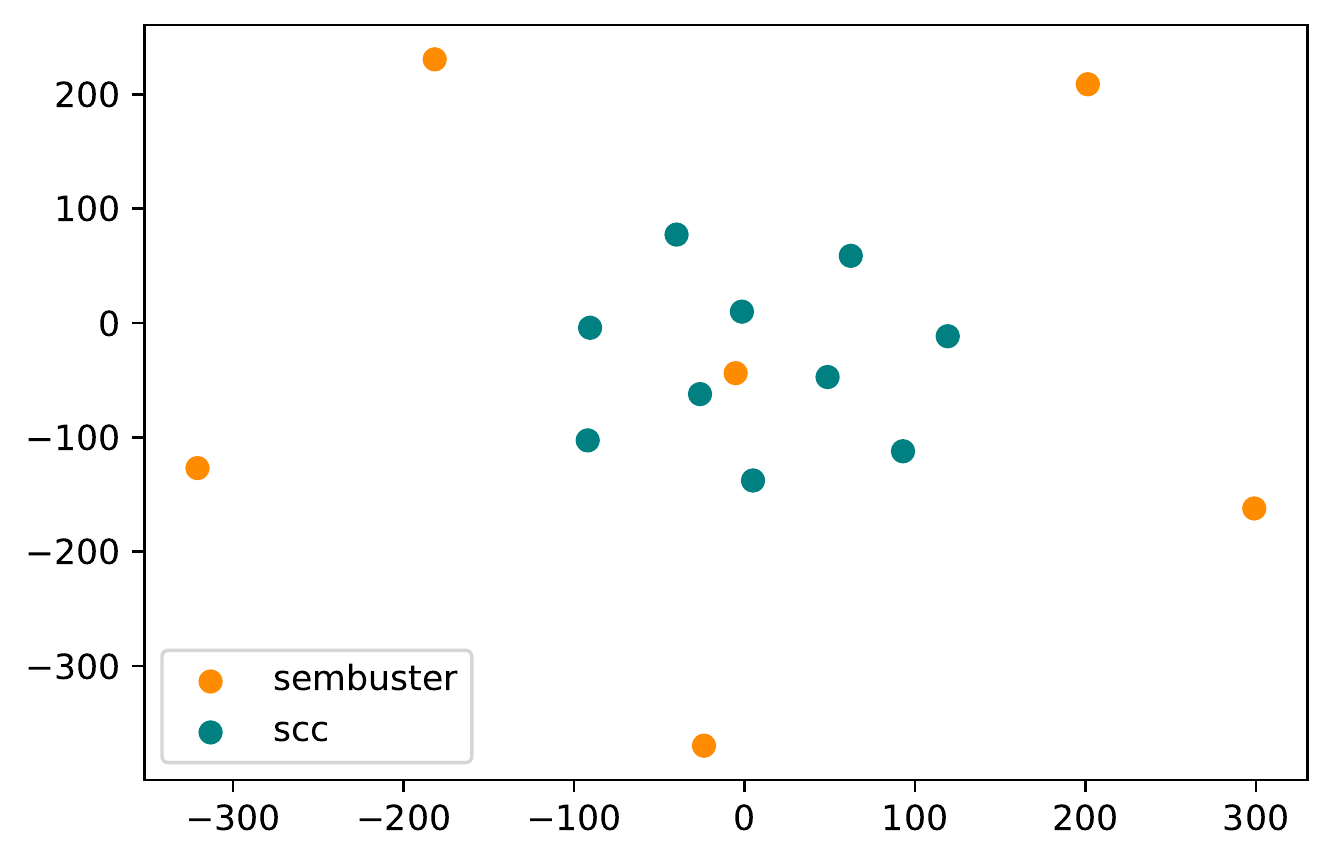}
	\end{subfigure}
	\begin{subfigure}[b]{0.45\linewidth}
		\includegraphics[width=\linewidth]{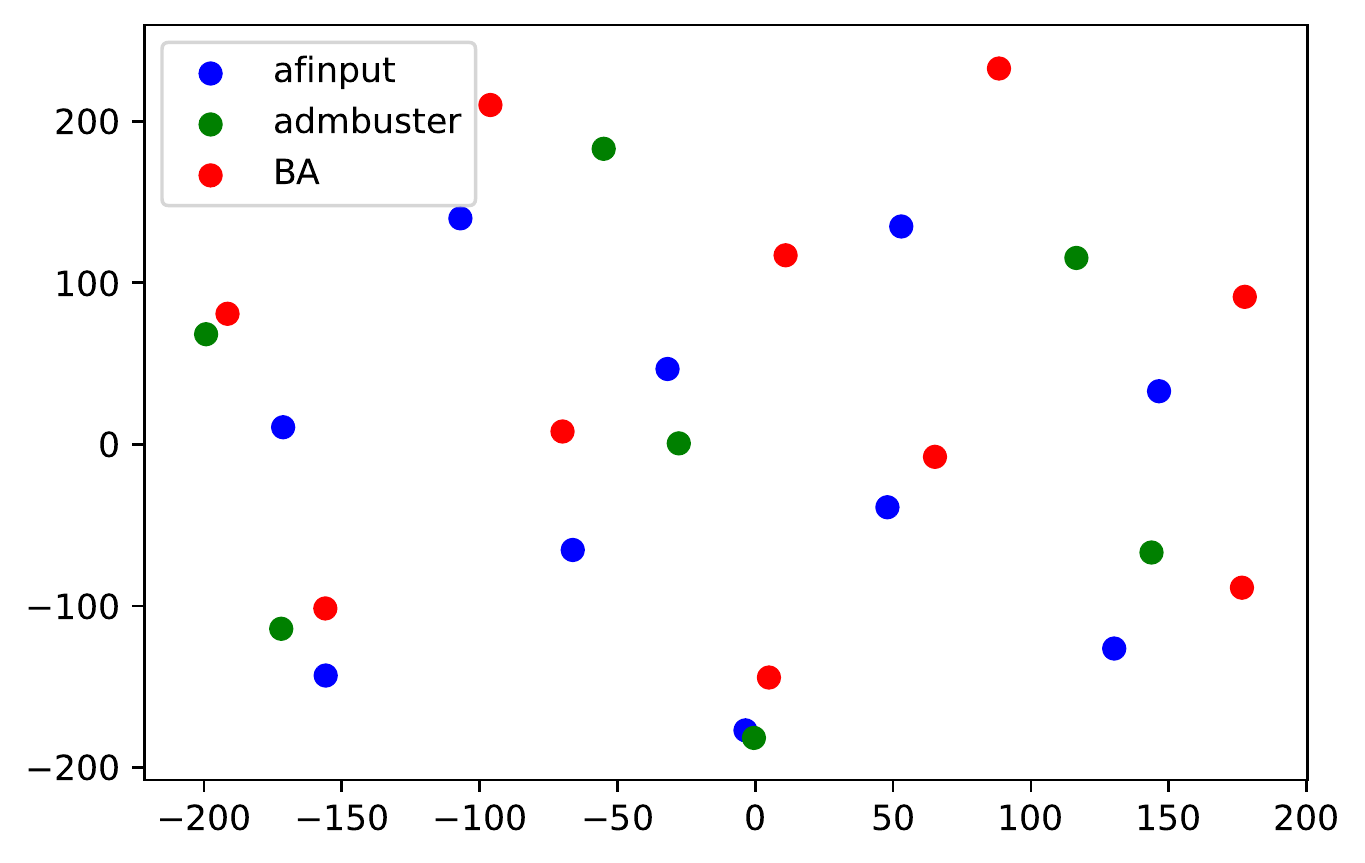}
	\end{subfigure}
	\begin{subfigure}[b]{0.45\linewidth}
		\includegraphics[width=\linewidth]{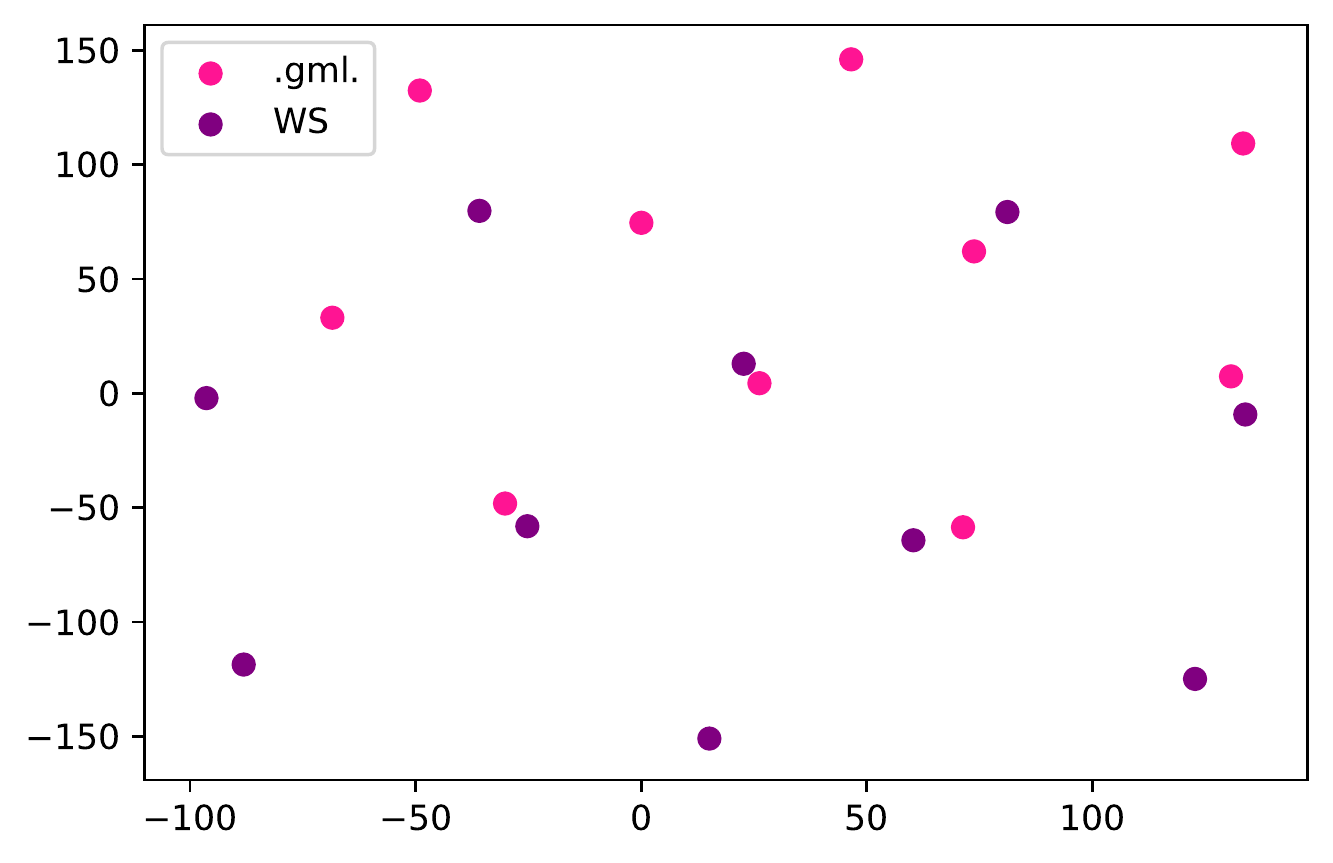}
	\end{subfigure}

	\caption{Graph-level Visualisation using t-SNE.}
	\label{fig:graphlevel}
\end{figure}
This had more ambiguous results. For most domains it was possible to identify a clear separation between classes either by clustering or by equal spacing. Examples of these are shown in Figure~\ref{fig:graphlevel}. First, for the \textit{ER}, \textit{.cnf}, and \textit{grd} domains we can see a clear separation between the three classes through three spacing patterns that are distinct by class. For \textit{sembuster} and \textit{scc} this pattern is even more distinct. However, for the afinput, admbuster, and BA domains it is less clear with less separation between the classes. For .gml and WS there is some overlap showing that there is a separation between the classes, but that some graphs are difficult to distinguish.

While this approach does demonstrate the feasibility of training and visualising graph-level embeddings of argumentation graphs, it is some way from being practically useful in its present form. What it does demonstrate is that a properly trained unsupervised embedding will be able to accurately separate argumentation graphs based on a training task that will then be available for visualisation. That means potentially being able to cluster corpora of argumentation graphs based on a variety of training tasks in order to discover new ways of classifying and categorising the arguments. In this case a simple supervised training task is used to generate a separation into already known classes, but both other well-known graph embeddings or a custom designed embedding for the task at hand might give improved results.

\section{Conclusion}
In this paper, we have shown that it is possible to use graph embeddings combined with t-SNE to visualise properties of argumentation graphs at both the node and the graph level. The node-level experiments were more successful, showing three distinct clusters according to the function of the arguments in the graph. However, the graph level example did shown that visual clustering of argument graphs by using a graph embedding is a least possible, although it needs some refinement to be applicable in practice. Further development of both the training task and the network architecture should yield improved results.

Wider exploration of this space would require the analysis of argumentation graphs of diverging provenance across many different domains. This would also involve testing a wider range of graph embeddings and eventually exploring the creation of embeddings from scratch with the kind of properties that would make them particularly useful for visualising argumentation graphs. The aim would be to enable the clustering argumentation graphs in an unsupervised manner to enable visual analysis of similarity at the graph level and commensurately to be able to see visual clustering of the individual arguments in graphs to be able to visually group arguments within a graph by some measure of similarity generated by the graph embedding.

\bibliography{ArgVis2020}
\bibliographystyle{vancouver}
\end{document}